\pdfoutput=1

\documentclass[11pt]{article}

\usepackage[]{acl}

\usepackage{times}
\usepackage{latexsym}

\usepackage[T1]{fontenc}

\usepackage[utf8]{inputenc}

\usepackage{microtype}

\usepackage{inconsolata}

\usepackage{graphicx}
\usepackage{booktabs}
\usepackage{framed}
\usepackage{flushend}
\usepackage{footnote}
\makesavenoteenv{tabular}

\title{FaithBench: A Diverse Hallucination Benchmark for Summarization by Modern LLMs}

\author{
 \textbf{Forrest Sheng Bao\textsuperscript{1}},
 \textbf{Miaoran Li\textsuperscript{1,2}},
 \textbf{Renyi Qu\textsuperscript{1}},
 \textbf{Ge Luo\textsuperscript{1}},
 \textbf{Erana Wan\textsuperscript{3}},
 \textbf{Yujia Tang\textsuperscript{4}},
 \textbf{Weisi Fan\textsuperscript{2}},
 \\
 \textbf{Manveer Singh Tamber \textsuperscript{5}},
 \textbf{Suleman Kazi\textsuperscript{1}},
 \textbf{Vivek Sourabh\textsuperscript{1}},
 \textbf{Mike Qi\textsuperscript{6}},
 \textbf{RuiXuan Tu\textsuperscript{6,7}},
 \\
 \textbf{Chenyu Xu\textsuperscript{2}},
 \textbf{Matthew Gonzales\textsuperscript{1}},
 \textbf{Ofer Mendelevitch\textsuperscript{1}},
 \textbf{Amin Ahmad\textsuperscript{1}},
\\
\\
 \textsuperscript{1}Vectara, Inc. Palo Alto, CA
 \textsuperscript{2}Iowa State University, Ames, IA
 \\
 \textsuperscript{3}University of Southern California, Los Angeles, CA
 \textsuperscript{4}Entropy Technologies, Melbourne, Australia
 \\
 \textsuperscript{5}University of Waterloo, Waterloo, ON
 \textsuperscript{6}\url{Funix.io}, Iowa City, IA
 \\
 \textsuperscript{7}University of Wisconsin, Madison, WI
 \small{
   \textbf{Correspondence:} {\{forrest, miaoran, amin\}\@vectara.com}
 }
}

\begin{document}
\maketitle

\begin{abstract}
Summarization is one of the most common tasks performed by large language models (LLMs), especially in applications like Retrieval-Augmented Generation (RAG). However, existing evaluations of hallucinations in LLM-generated summaries, and evaluations of hallucination detection models both suffer from a lack of diversity and recency in the LLM and LLM families considered. This paper introduces FaithBench, a summarization hallucination benchmark comprising challenging hallucinations made by 10 modern LLMs from 8 different families, with ground truth annotations by human experts. ``Challenging'' here means summaries on which popular, state-of-the-art hallucination detection models, including GPT-4o-as-a-judge, disagreed on. Our results show GPT-4o and GPT-3.5-Turbo produce the least hallucinations. However, even the best hallucination detection models have near 50\% accuracies on FaithBench, indicating lots of room for future improvement.
\end{abstract}

\section{Introduction}
With the increasing use of Large Language Models (LLMs) to process textual data, ensuring their trustworthiness has become a critical concern. In applications such as Retrieval Augmented Generation (RAG)~\cite{facebook_meta_rag_nips_2020}, LLMs are used to generate answers or summaries from textual input. 
When the generated text includes unsupported information, it is considered a hallucination, which can be misleading or harmful.

Understanding the state of hallucinations in LLMs is crucial but hard. 
Existing hallucination leaderboards, such as Vectara's Hallucination Leaderboard~\footnote{\url{https://huggingface.co/spaces/vectara/leaderboard}} and Galileo's Hallucination Index~\footnote{\url{https://www.rungalileo.io/hallucinationindex}}, detect hallucinations using models such as Google's TrueTeacher~\cite{gekhman-etal-2023-trueteacher}, Vectara's HHEM-2.1-Open~\cite{hhem-2.1-open}, or even GPT series models in a zero-shot, LLM-as-a-judge fashion~\cite{luo2023chatgptfactualinconsistencyevaluator,Geval}. These detection models are known to have an accuracy below 80\% on benchmarks such as AggreFact~\cite{tang-etal-2023-understanding} and RAGTruth~\cite{niu-etal-2024-ragtruth}. Moreover, existing benchmarks often rely on a narrow selection of LLMs, many of which are outdated and lack diversity across model families.
If we assume LLMs hallucinate differently---due to variations in training methods, datasets, and architectures, as well as changes in behavior as models scale up---then conclusions drawn from such benchmarks are incomplete, capturing only specific types of hallucinations.

To address this gap, the industry and research community need a hallucination benchmark that includes modern LLMs across diverse model families, along with human-annotated ground truth for more reliable evaluation. This paper presents FaithBench, a summarization hallucination benchmark built on top of Vectara's Hallucination Leaderboard which is popular in the community~\cite{hugginface-hallucinations-leaderboard,merrer2024llms} because it contains summaries generated by dozens of modern LLMs.
We add human annotations, including justifications at the level of individual text spans, to summaries from 10 LLMs belonging to 8 LLM families. 
To make the best use of our annotators' time, we focus on labeling challenging samples where hallucination detectors disagree the most, as obvious hallucinations can be reliably detected automatically. 
The majority of our annotators are experts in the field of hallucination detection, with half of them having published hallucination-related papers at major NLP conferences.

FaithBench allows us to evaluate both the hallucination rates of LLMs and the accuracy of hallucination detection models.
To the best of our knowledge, this is the first evaluation of hallucinations across 10 LLMs and 8 LLM families using human-annotated ground truth.
GPT-4o has the lowest hallucination rate, followed by GPT-3.5-Turbo, Gemini-1.5-Flash, and Llama-3-70B.
All hallucination detectors are found to correlate poorly with human-annotated ground truth, with the best F1-macro score and balanced accuracy at 55\% and 58\% respectively. This highlights our limited understanding of hallucinations and the challenges ahead.

We hope that FaithBench can catalyze research into detecting and mitigating hallucinations in LLMs. In contrast with existing benchmarks, FaithBench 1) covers a wide array of LLM families and diverse hallucination characteristics, 2) factors the subjectivity of hallucination perception, by expanding binary consistent vs. unfaithful labels to include two new ``gray-area'' labels: ``questionable'' and ``benign'', 3) includes only challenging hallucination samples. The  repo is \url{https://github.com/vectara/FaithBench}


    

\section{The Benchmark}
\subsection{Definition of hallucinations}

The word ``hallucinating'' has two meanings in the context of LLMs. 
It could mean either ``non-factual''~\cite{mishra2024finegrainedhallucinationdetectionediting, ji-etal-2024-anah, ji-etal-2023-towards, deng2024pfmemodularapproachfinegrained, li-etal-2024-dawn, chen2023felm}, when the LLM-generated text is not supported by the world knowledge, or ``unfaithful'' or ``inconsistent''~\cite{tang-etal-2023-understanding, niu-etal-2024-ragtruth, tang-etal-2024-tofueval} when the LLM-generated text does not adhere to its input.
This paper focuses on the latter case, wherein an LLM is expected to fulfill a task, often generating a summary or answering a question, based on a given passage or reference. Such scenarios are common in applications such as Retrieval-Augmented Generation (RAG)~\cite{facebook_meta_rag_nips_2020}.
By this definition, a statement can be simultaneously factual yet unfaithful. For example, if the passage states that ``water has a smell'', then the statement ``water is odorless'' is a hallucination despite being factual according to common world knowledge.

\subsection{Hallucination Taxonomy}
\label{sec:taxonomy}

While hallucinations draw a great deal of attention in NLP because they are so often harmful and misleading, recent research argues that not all hallucinations are necessarily bad~\cite{circumstantial_inference}. 
In fact, users often value the enrichment LLMs provide through reasoning, creativity, and factual knowledge.
Hence, we separate hallucinations into \emph{benign} and \emph{unwanted} categories.

Given that some hallucinations are disputed even among human annotators, this paper categorizes hallucinations into three types:
\begin{itemize}
  \item \textbf{Questionable}: not clearly a hallucination, classification may differ depending on whom you ask. 
  \item \textbf{Benign}: clearly a hallucination, but supported by world knowledge, common sense, or logical reasoning, such that a reader finds it acceptable or welcomed. 
  \item \textbf{Unwanted}: A clear hallucination that is not benign. This category is further subdivided into two categories:
  \begin{itemize}
    \item \textbf{Intrinsic}: Contradicted by the passage, either in part or in whole.
    \item \textbf{Extrinsic}: neither supported by the passage, nor inferrable from it, nor factual.
  \end{itemize}
\end{itemize}

\subsection{Data Sampling}
\paragraph{Sourcing the data}
We utilize Vectara's hallucination leaderboard, which already contains summaries generated by dozens of LLMs and is frequently cited in the community.
In the leaderboard dataset, the passages for summarization come from various Natural Language Inference (NLI), fact-checking, or summarization datasets. Some passages are specifically crafted to `trick' LLMs into hallucinating (Appendix~\ref{sec:app:trick}), such as by combining information about two unrelated individuals in the same profession within one passage to induce a coreference error.
A \emph{sample} is defined as a pair consisting of a source passage and an LLM-generated summary.

\paragraph{Filtering samples by LLM} 
To balance annotator effort with our goal of LLM diversity, we restrict the benchmark to eight of the most anecdotally popular LLM families: GPT, Llama, Gemini, Mistral, Phi, Claude, Command-R, and Qwen.
For each family, we then selected the smallest version in its latest generation. The exceptions are the GPT and Llama series from which we select two each. For GPT, we select GPT-4o and GPT-3.5-Turbo as they are cost efficient. For Llama, we select Llama-3.1-70B and -8B in order to assess the impact of model size.
Our preference towards small and affordable models aims to maximize the value of our work to the community as these models are used more widely than their larger counterparts.

\paragraph{Filtering samples by consensus of detectors}
Human annotation of obvious hallucinations is of limited value, as they can be easily detected by automatic systems; the real value lies in annotating challenging samples where popular detection models disagree.
This will provide a valuable calibration for the community, highlighting areas where detectors struggle and guiding future improvements. 
Based on their popularity~\cite{shroom,grapheval}, the following hallucination detectors are chosen to identify challenging samples: Google's True-NLI~\cite{honovich-etal-2022-true-evaluating} and TrueTeacher~\cite{gekhman-etal-2023-trueteacher}, Vectara's HHEM-2.1-Open~\cite{hhem-2.1-open}, and GPT-\{4o, 3.5-Turbo\}-as-a-judge~\cite{Geval,luo2023chatgptfactualinconsistencyevaluator}. 

\paragraph{Sample groups}

In this paper, our samples are divided into groups of ten which share one common source passage but contain outputs from 10 different LLMs. This allows us to compare the performance of each LLM while controlling for the characteristics of the source text. 

We then rank groups by the number of challenging summaries in each group.
The top 115 groups containing at least 7 challenging summaries each are moved to the next step. 

\subsection{Human Annotation}

\paragraph{Annotators}
The hallucination ground truth is added by 11 human annotators. 
The super majority of them are experts in the field of hallucination detection, with half of them having published hallucination-related papers at top-tier NLP conferences.
About half of them are graduate students from three US/Canadian universities, and the other half are machine learning engineers. 
The diverse yet professional backgrounds of the annotators helps to ensure the quality of the annotations.
Three annotators are native speakers of English. 
All annotators are aware that the data they created will be made open source to the public.

\paragraph{The pilot run}

A pilot run of 30 random samples pertaining to 30 different passages was conducted to ensure annotators are in agreement on the definition and categorization of hallucinations.

The pilot run revealed two issues. First, many sports-related samples required specific knowledge of European sports terminology, which posed a challenge for our annotators who are not familiar with these sports.
Second, many source passages are not self-consistent due to noise introduced in their construction.
Based on these observations, we visually inspected all passages and removed corresponding samples, leaving us with 800 samples. 

The samples were then divided into 16 batches of 50 samples each (5 passages $\times$ 10 LLM-generated summaries). All batches were annotated by two annotators with most also having a third annotator to provide an additional opinion. In the process of post-pilot annotation, we found more samples with noisy passages including image captions or advertisements. They are then excluded from the benchmark. The final benchmark totals at 660 samples (66 passages $\times$ 10 LLMs).

\paragraph{Semantic-assisted cross-checking}
Given a text span in the summary, finding corresponding spans in the passage that support or refute it is often difficult because modern LLMs are very abstractive, limiting the benefit of exact string matching.
Thus, we developed an in-browser annotation tool that highlights sentences in the passage that are semantically similar to a selected text span in the summary. 
With the benefit of this annotation tool, annotators are asked to select all spans in the summary that are hallucinations or suspected hallucinations. For each selected span, they are asked to assign a label (\S~\ref{sec:taxonomy}) and add a note explaining their reasoning. If the span is related to one in the passage, they are encouraged to link the summary span and the passage span. 

\section{Results}
\subsection{Annotation quality}
\label{sec:alpha}
Following the common practices in the field, the annotation quality is measured by inter-annotator agreement (IAA) using 
Krippendorff's alpha~\cite{krippendorff2018content} at the sample level. 

Different spans in a summary maybe assigned different labels by the same annotator. 
To compute IAA, each sample's span-level labels are ``worst-pooled'' into one sample-level label using the worst label among all spans assigned by the annotator. The severity of hallucinations is ordered as: consistent (best) $\succ$ benign $\succ$ questionable $\succ$ unwanted (worst). 

The IAA for the ``consistent'' and ``unwanted'' classes is 0.748. 
Undoubtedly, the IAA for the other two classes, ``questionable'' and ``benign'', will be low. 
The IAA for binary classification consistent + benign vs. unwanted, and ternary classification consistent + benign vs. questionable vs. unwanted, are 0.679 and 0.58, respectively. The much lower IAA after considering the ``questionable'' and ``benign'' labels indicates the high subjectivity on borderline hallucinations and justifies the necessity of introducing them in our benchmark. 

Annotations are done in two rounds. In the first round, annotators work independently. In the second round, they discuss and resolve disagreements. Annotators are encouraged to hold their ground if they are confident in their annotations rather than being forced to converge with other annotators.
IAA for the first round can be as low as 0 while the second round significantly boost the IAA. This reflects the challenge in annotating hallucinations that even experience professionals can miss them. 


\subsection{Ranking LLMs by Hallucinations}

\begin{figure}[!htbp]
    \centering
    \includegraphics[width=0.95\columnwidth]{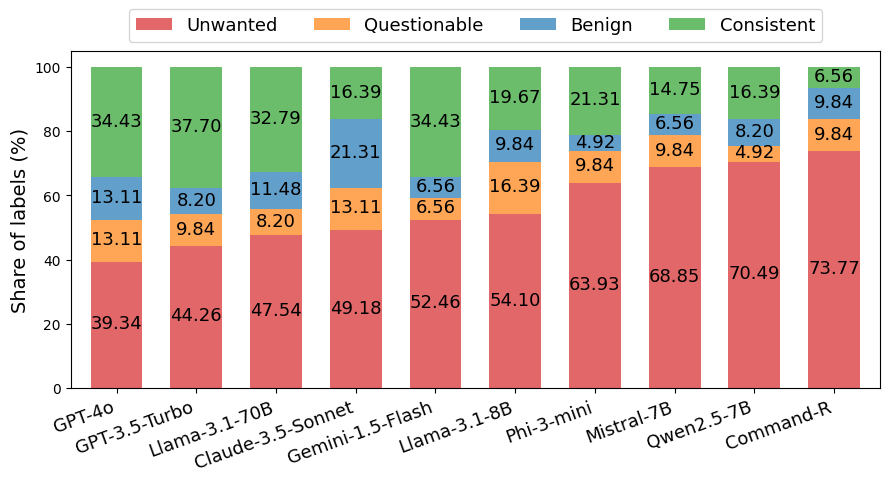}
    \caption{Distribution of samples by the most severe hallucination label per LLM.}
    \label{fig:label_dist}
\end{figure}

Figure~\ref{fig:label_dist} shows the distribution of 
``worst-pooled'' (\S~\ref{sec:alpha}), sample-level  labels per LLM. 
GPT-3.5-Turbo produces the highest percentage (37.70\%) of fully consistent summaries. 
GPT-4o and Gemini-1.5-Flash tie the No.2 spot, with nearly 1/3 of the summaries produced by them are fully consistent. Claude-3.5-Sonnet produces a great amount (21.31\%) of summaries that contain benign hallucinations. 

Using the sample-level labels, we can compute the rate of hallucinations of LLMs and rank them (Table~\ref{tab:llm-hallucination-rates}). 
The rankings according to FaithBench (first three columns) generally align well with the ranking in Vectara's Hallucination Leaderboard (rightmost column). It slightly differs from Galileo's Hallucination Index, which ranks Claude-3.5-Sonnet as the best proprietary LLM. 

\begin{table}[!htbp]
  \centering

  \scriptsize
  \begin{tabular}{l|cccc}
    \toprule
    \textbf{LLM} & \textbf{Unwanted} & \textbf{U+Q} & \textbf{U+Q+B} & \textbf{VHL} \\
    \midrule
    GPT-4o & 39.34 (1) & 52.46 (1) & 65.57 (2) & 1 \\
    GPT-3.5-Turbo & 44.26 (2) & 54.1 (2) & 62.3 (1) & 2 \\
    Llama-3.1-70B & 47.54 (3) & 55.74 (3) & 67.21 (4) & 3 \\
    Gemini-1.5-Flash & 52.46 (5) & 59.02 (4) & 65.57 (2) & 4 \\
    Llama-3.1-8B & 54.1 (6) & 70.49 (6) & 80.33 (6) & 5 \\
    Claude-3.5-Sonnet & 49.18 (4) & 62.3 (5) & 83.61 (7) & 6 \\
    Qwen2.5-7B & 70.49 (9) & 75.41 (8) & 83.61 (7) & 7 \\
    Phi-3-mini-4k & 63.93 (7) & 73.77 (7) & 78.69 (5) & 8 \\
    Command-R & 73.77 (10) & 83.61 (10) & 93.44 (10) & 9 \\
    Mistral-7B & 68.85 (8) & 78.69 (9) & 85.25 (9) & 10 \\
    \bottomrule
  \end{tabular}  
  \caption{Hallucination rates (\%) and LLM rankings (between parenthesis) based on three levels: Unwanted only (U), U + Questionable (U+Q), and U+Q+Benign (U+Q+B). Column VHL is the ranking of LLMs in Vectara's Hallucination Leaderboard.}
  \label{tab:llm-hallucination-rates}
\end{table}

\begin{figure}[!htbp]
    \centering
    \includegraphics[width=\linewidth]{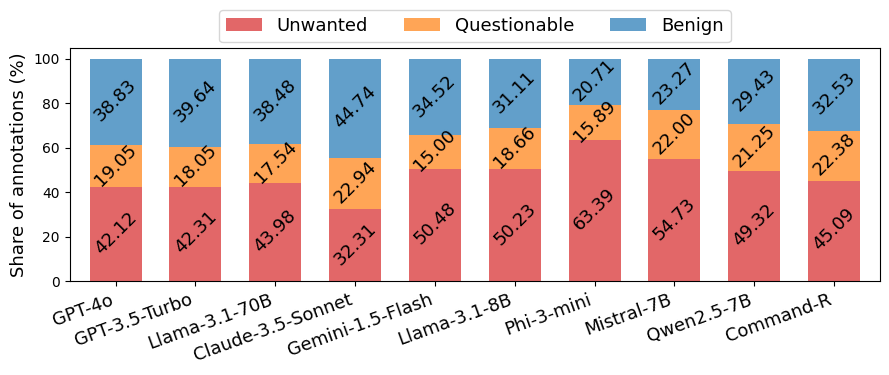}
    \caption{Distribution of hallucination annotations.}
    \label{fig:halu_dist}
\end{figure}

Figure~\ref{fig:halu_dist} presents, for each LLM,  
the ratios of unwated, questionable, and benign annotations (span-level) to all hallucination annotations.
When interpreting all results above, it is important to keep in mind that they are only true for the challenging samples. It may not be true for all samples.

\subsection{Ranking Hallucination Detectors}

Table~\ref{tab:detector-performance} shows the balanced accuracy (BA) and F1-Macro score of several hallucination detectors against the ground truth in FaithBench.
Here a sample is hallucinated if it is unwanted or questionable. 
The balanced accuracies of all detectors are near 50\%, 
indicating the rigor of FaithBench and the need for a challenging benchmark like FaithBench in our battle against hallucinations. 
For zero-shot usage of GPT's, we use the prompt template in~\cite{luo2023chatgptfactualinconsistencyevaluator}.

\begin{table}[!htbp]
\scriptsize 
  \centering
  \begin{tabular}{p{3.5cm}|ccc}
    \toprule
    \textbf{Hallucination Detector} & \textbf{BA  (\%)} & \textbf{F1-Macro  (\%)} \\
    \midrule
    HHEM-2.1~\cite{hhem_2.1_trilingual} & 55.68 & 40.86 \\
    HHEM-2.1-Open~\cite{hhem-2.1-open} & 51.37 & 32.40 \\
    HHEM-1 & 48.96 & 41.63 \\
    \midrule
    True-Teacher~\cite{gekhman-etal-2023-trueteacher} & 54.21 & 39.21 \\
    True-NLI~\cite{honovich-etal-2022-true-evaluating} & 50.62 & 28.17 \\
    \midrule
    GPT-4-Turbo, zero-shot & 57.65 & 43.61 \\
    GPT-4o, zero-shot & 56.29 & 40.75 \\    
    GPT-4, zero-shot & 53.45 & 33.54 \\
    GPT-3.5-Turbo, zero-shot & 44.91 & 37.41 \\
    \midrule
    MiniCheck-Roberta-L~\cite{tang-etal-2024-minicheck} & 55.03 & 53.35 \\
    MiniCheck-Deberta-L & 54.95 & 54.90 \\
    MiniCheck-Flan-T5-L & 50.50 & 49.52 \\
    \bottomrule
  \end{tabular}  
  \caption{Performance of hallucination detectors. }
  \label{tab:detector-performance}
\end{table}


\section{Conclusion}

This paper introduces FaithBench, a benchmark for summarization hallucinations, featuring human-annotated hallucinations in summaries generated by 10 modern LLMs across 8 different model families.
To account for the subjective nature of hallucination perception, we introduced two gray-area labels---\emph{questionable} and \emph{benign}---in addition to the common binary labels of \emph{consistent} and \emph{hallucinated}.
The human annotation is fine-grained at the span level and most annotations are accompanied by reasons for better explainability. 
With FaithBench, we are able to rank the state-of-the-art LLMs and hallucination detectors. 
While the ranking of LLMs largely aligns with a popular hallucination leaderboard, hallucination detectors only achieve around 50\% accuracy on FaithBench. 
In summary, the creation and curation of FaithBench mark a crucial step in the long journey towards effectively addressing hallucinations.


\section*{Limitations} 

Although a primary goal of FaithBench is the diversity of hallucinations in various characteristics, as a short paper, it cannot cover a lot. 

FaithBench covers only summarization. 
There are many other tasks where hallucination detection is needed such as question answer. 

Due to the composition of the foundation dataset, most passages are between 106 (1st quartile) to 380 
(3rd quartile) English words in length (Appendix~\ref{sec:app:data_details}). This translates to roughly 137 to 494 tokens. This means that FaithBench only measure short-context hallucinations for LLMs. We will extend it to include samples of longer contexts, such as using those in RAGTruth as the passages. 
But that will raise the human annotation difficulties and cost. 

Due to the tremendous amount of labor needed in human annotation, we are not able to cover models of various sizes in the same family. 
This limits our ability to study the impact model sizes in hallucination. 

The spans and reasoning collected in FaithBench are not used in evaluating LLMs and hallucination detectors. 

Because FaithBench only contains challenging samples, 
our ranking to LLMs and hallucination detectors does not reflect their rankings on all samples. 
When interpreting all results above, it is important to keep this in mind.

Lastly, although FaithBench 
makes the effort to factor in subjectivity in labeling questionable and benign hallucinations, 
the inter-annotator agreements on the two gray-area hallucinations are low.
We will need to develop a better taxonomy of hallucinations after taking a closer look such annotations/samples.

\bibliography{custom}

\appendix

\section{Hallucinations vs. lengths}

Here we study the relationship between hallucinations and passage length. 
When interpreting the results, 
please factor in the length distribution of passages (Appendix~\ref{sec:app:data_details}). 
Points beyond 400 words are covered very sparsely. 

\begin{figure}[!htbp]
  \centering
  \includegraphics[width=\columnwidth]{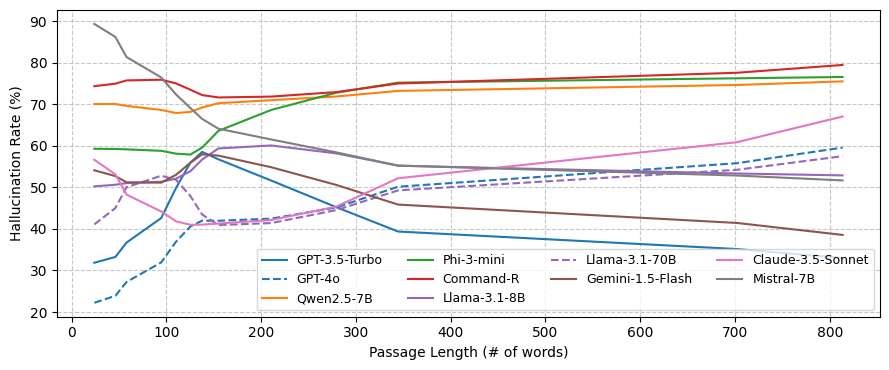}
  \caption{Hallucination rates vs. passage length}
  \label{fig:hallucination-rate-vs-passage-length}
\end{figure}

Figure~\ref{fig:hallucination-rate-vs-passage-length} shows the relationship between hallucination rates (considering only unwanted hallucinations) and the length of the passage. 
Contrary to the expectation that longer passages lead to more hallucinations, some models exhibit higher hallucination rates with shorter passages. Upon examining randomly sampled hallucinations for short passages, we found that LLMs often add extra information not present in the source, which is also difficult to validate even with external knowledge.


\begin{figure}[!htbp]
  \centering
  \includegraphics[width=\columnwidth]{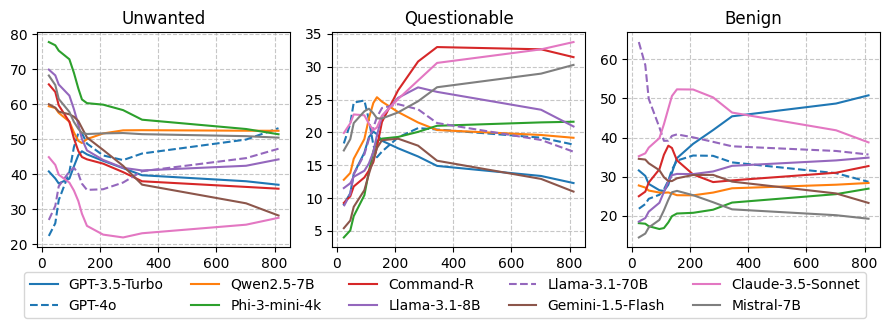}
  \caption{Ratio (\%) of hallucination vs. passage length}
  \label{fig:hallucination-ratio-vs-passage-length}
\end{figure}

We further study the percentage of hallucination types relative to source passage length. 
As shown in Figure~\ref{fig:hallucination-ratio-vs-passage-length}, most LLMs exhibit a decrease in the ratio of unwanted hallucinations as the passage length increases. The ratios of questionable and benign hallucinations show mixed trends across models, indicating that the relationship between hallucination types and passage length is inconsistent and model-specific.

\begin{figure}[!htbp]
  \centering
  \includegraphics[width=\columnwidth]{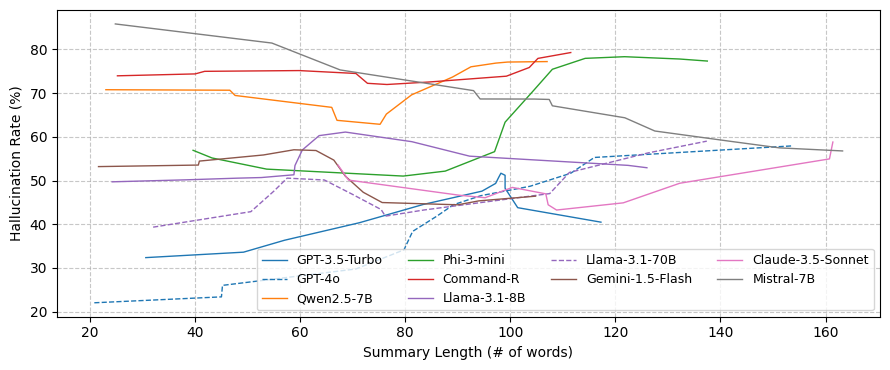}
  \caption{Hallucination rates vs. summary length}
  \label{fig:hallucination-rate-vs-summary-length}
\end{figure}

Studying the relationship between the hallucination rates and the length of the summary is a bit hard because different LLMs yield summaries of different lengths. 
Despite that, we manage to get Figure~\ref{fig:hallucination-rate-vs-summary-length}.

\section{Data Source details}
\label{sec:app:data_details}

The mean, median, and standard deviation of the lengths of passages are 300, 184, and 277 respectively. 
The 1st, 2nd, 3rd, and 4th 5-quantiles of passage lengths fall onto 87, 133, 282, 593 words.

Composition of Vectara's Hallucination Leaderboard is given in Table~\ref{tab:leaderboard_mix}. Some samples are created with the intention to trick LLMs into hallucinating.

\begin{table}[!htbp]
    \centering
    
    \scriptsize
    \begin{tabular}{lr}
\toprule
 dataset &  Percentage \\
\midrule
XSum-Factuality~\cite{maynez_acl20} & 27.34 \\
FEVER, dev \cite{Thorne18Fever} & 25.85 \\
Polytope, test \cite{laban-etal-2022-summac} & 18.79 \\
VitaminC, dev \cite{vitaminc} & 11.23 \\
SummEval, valid \cite{fabbri2020summeval} & 9.94 \\
Frank, valid \cite{pagnoni-etal-2021-understanding} & 6.86 \\
\bottomrule
\end{tabular}
\caption{Composition of Vectara's Hallucination Leaderboard}
    \label{tab:leaderboard_mix}
\end{table}



\section{Annotator instructions and the annotation tool}

\paragraph{Instruction to Annotators\\}

The task is to label how faithful the output of an LLMs is to the input given to it. 

In a RAG system, text retrieved based on a user query is called the “context”. The context forms part of the input to an LLM to produce a summary that answers the user query. 

Please select any text span in the summary that is not faithful to or supported by the context, and categorize it to one or multiple types of hallucination. If there is any text span in the context that is related to the summary span, please select it and link it with the summary span. 

A faithful response can be contradictory to the world or your knowledge as long as such knowledge is in the context too. Do not confuse “faithful” with “factual”.  

\{\{Hallucination Taxonomy \}\} 

\{\{Hallucination Examples \}\}

\begin{figure*}[!htbp]
    \centering
    \includegraphics[width=\linewidth]{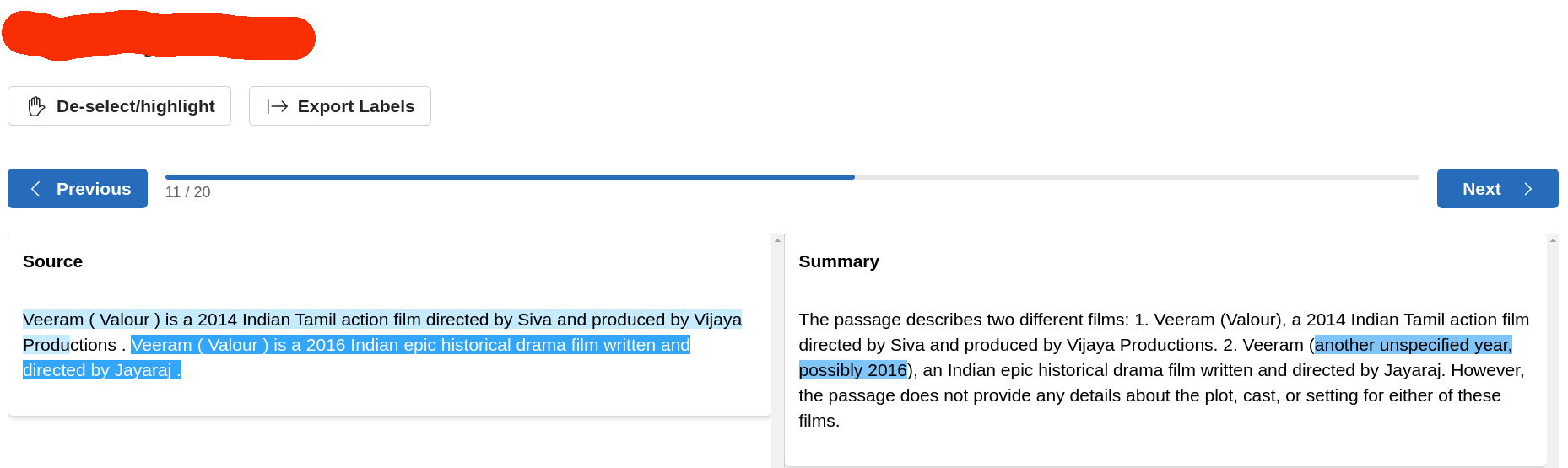}
    \caption{Semantic highlighting for easy cross-checking in our annotation tool. The selected summary span is embedded when selected. Then its dot-product distance to sentences, whose embeddings are precomputed during ingestion, in the passage are computed. Finally, sentences in the passage are highlighted with different color intensity proportional to their semantic distances. }
    \label{fig:semantic_highlight}
\end{figure*}

\begin{figure*}
    \centering
    \includegraphics[width=\linewidth]{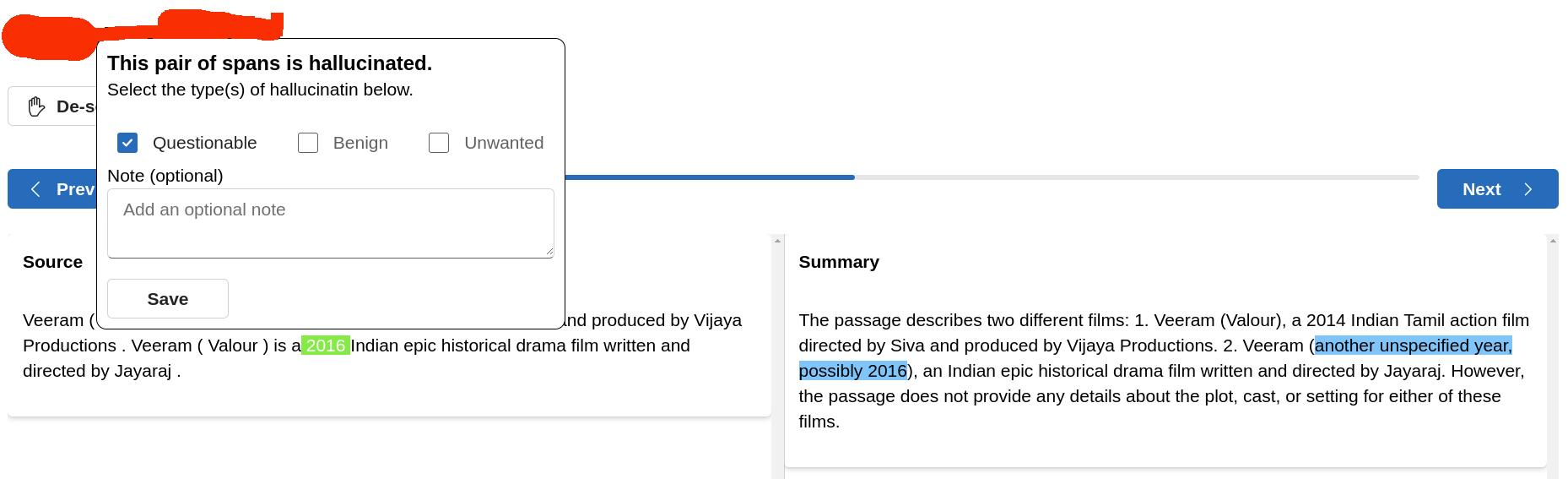}
    \caption{Annotating a pair of selected spans. }
    \label{fig:tool_annotate}
\end{figure*}

\paragraph{Annotation tool}
The semantic cross-checking feature of our annotation tool is given in Figure~\ref{fig:semantic_highlight}. 
Figure~\ref{fig:tool_annotate} shows that a pair of text spans, one in the passage and the other in the summary, are selected and their labels are being added in the pop-up bubble. 

\section{Hallucination Taxonomy and examples}
Short examples are: 
\begin{itemize}
    \item Questionable
    \begin{itemize}
        \item Last August \\ -->  the August of last year 
        \item The train was late by 2 hours 45 minutes \\ --> The train was late by almost 3 hours. 
    \end{itemize}
    \item Benign
    \begin{itemize}
        \item I ate a lot for lunch. \\ -->
 Overeating causes obesity. 
        \item Tesla’s Model S is sold for \$79k.  \\ --> Model S is made by Tesla. \\
        \textcolor{blue}{(Common sense tells us that Tesla is not a person but a manufacturer here.)}
        \item President Biden visited Japan today \\ --> Joe Biden was in Japan today. \\
        \textcolor{blue}{(The first name of Biden is not mentioned in the passage. But we Chauvinistically assume that most people in the world know the first name of the current US president.)}
        \item At the University of Mississippi, about 55 percent of its undergraduates and 60 percent overall come from Mississippi, and 23 percent are minorities; international students come from 90 nations \\ -->  
        The University of Mississippi has a diverse student body. \\
        \textcolor{blue}{(This is hallucination because the passage does not assess diversity. But it is reasonable to infer. Hence, benign hallucination.)}
    \end{itemize}
    \item Unwanted
    \begin{itemize}
        \item I ordered a pizza from downstairs.\\ --> The pizza is yummy. \\
        \textcolor{blue}{(This is an extrinsic hallucination.)}
        \item I ate the pizza \\ -->  I tossed away the pizza. \\
        \textcolor{blue}{(This is an intrinsic hallucination because the summary cannot be true when the passage is also true.)}
        \item  Goldfish weigh 1 pound and can grow up to 30 cm while koi weigh up to 2 pounds and are as long as 2 meters.  \\ --> 
Koi weigh 1 pound and can grow up to 2 meters.  \\
        \textcolor{blue}{(This kind of hallucinations are often referred to as discourse hallucinations where pieces of information are stitched together wrongly.)}  \item The Earth was believed flat. \\  --> The Earth was flat. 
        \item Penguins cannot fly. \\ -->  No birds can fly. 
        \item Company X employees 50,000 people \\ -->  Company Y employees 50,000 programmers. 

    \end{itemize}
\end{itemize}

Long examples are shown in Figure~\ref{fig:app:examples}.

\begin{figure*}[b]
\fboxsep=5pt 
\fboxrule=0.8pt 

\fbox{ 
\begin{minipage}{\textwidth} %
\small
\textbf{Source:} The incident occurred on the A9 north of Berriedale in Caithness at about 14:00. Several other people received minor injuries in the crash. The road was closed and diversions were put in place via the A897 Helmsdale to Melvich road. \textcolor{blue}{Police Scotland said inquiries into the incident were ongoing.}\\

\textbf{Summary:} The passage describes an incident that occurred on the A9 north of Berriedale in Caithness at around 14:00. This incident resulted in a crash, and several people received minor injuries. The road was closed, and diversions were set up via the A897 Helmsdale to Melvich road. \textcolor{blue}{Police Scotland is currently conducting ongoing inquiries into the incident.}\\

\textbf{Explanation:} The source describes a past event, but "is currently conducting" in the summary implies present action, causing confusion.\\

\textbf{Human Label:} Questionable\\

--------------------------------------------------------

\vspace{0.3cm}
\textbf{Source:} Sheryl Lee. \textcolor{blue}{In 2016, she appeared in Cafe Society, and also completed the Showtime revival of Twin Peaks (2017), reprising her role of Laura Palmer.} Sheryl Lee Ralph. In 2016 she played Madame Morrible in Wicked on Broadway.\\

\textbf{Summary:} Sheryl Lee appeared in the film Cafe Society in 2016 and \textcolor{blue}{reprised her role as Laura Palmer in the Showtime revival of Twin Peaks in 2017}. Sheryl Lee Ralph played Madame Morrible in the Broadway production of Wicked in 2016. \\

\textbf{Explanation:} The source indicates that 2017 was the release year for the Showtime revival of Twin Peaks, while Sheryl Lee completed her role in 2016. The summary misrepresents this timeline, causing a distortion of the original information.\\

\textbf{Human Label:} Unwanted.Intrinsic\\

--------------------------------------------------------

\vspace{0.3cm}
\textbf{Source:} As of 22 February 2020 , 77,984 cases have been confirmed , including in all provinces of China and more than two dozen other countries .\\

\textbf{LLM Summary:} As of 22 February 2020, there were 77,984 confirmed cases \textcolor{blue}{of the disease}, present in all Chinese provinces and over two dozen other countries.\\

\textbf{Explanation:} The source does not specify that the cases are related to a disease, and this information cannot be inferred with certainty.\\

\textbf{Human Label:} Unwanted.Extrinsic\\

--------------------------------------------------------

\vspace{0.3cm}
\textbf{Source:} Albert Lawrence Brooks (born Albert Lawrence Einstein; July 22, 1947) is an American actor, filmmaker, author and comedian. Finding Dory. Directed by Andrew Stanton with co-direction by Angus MacLane, the screenplay was written by Stanton and Victoria Strouse.\\

\textbf{Summary:} The passage provides information about Albert Lawrence Brooks, an American actor, filmmaker, author, and comedian. It also mentions \textcolor{blue}{the film ``Finding Dory,''} which was directed by Andrew Stanton with co-direction by Angus MacLane. The screenplay for the film was written by Stanton and Victoria Strouse.\\

\textbf{Explanation:} The source does not explicitly state that ``Finding Dory'' is a film, but this can be reasonably inferred from the context or verified with general knowledge.\\

\textbf{Human Label:} Benign\\

\end{minipage}}
\caption{Examples of each hallucination type}
\label{fig:app:examples}
\end{figure*}


\section{More related work}

Table~\ref{tab:benchmark_coverage} shows the LLM families covered by different benchmarks. In all benchmarks, GPT family is covered. Llama models are also widely explored, covered in 5 benchmarks. Many of the benchmarks in Table~\ref{tab:benchmark_coverage} are for factuality rather than faithfulness in this paper, or do have have human ground truth.

\begin{table*}[!htbp]
  \small
  \centering

    \begin{tabular}{c|c}
    \toprule
    \textbf{Benchmark} & \textbf{
Model Families}\\
    \midrule
    FELM~\cite{chen2023felm} & GPT\\
    FactCHD~\cite{ijcai2024p0687} & GPT\\    FavaBench~\cite{mishra2024finegrainedhallucinationdetectionediting} & Alpacab, Vicuna, Llama2\\
    ANAH~\cite{ji-etal-2024-anah} & GPT, InternLM\\
    RAGTruth~\cite{niu-etal-2024-ragtruth} & GPT, Mistral, Llama\\
    TofuEval~\cite{tang-etal-2024-tofueval} & GPT, Vicuna, WizardLM \\
    HaluEval-2.0~\cite{li-etal-2024-dawn} &GPT, Claude, Vicuna , YuLan, Llama-2, Alpaca \\    HalluDial~\cite{luo2024halludiallargescalebenchmarkautomatic} & Mistral, Vicuna, Llama-2, and GPT \\    WildHallucination~\cite{zhao2024wildhallucinationsevaluatinglongformfactuality} & GPT, Gemini, Claude, Command R, Perplexity, Llama, Mistral/Mixtral\\
    \bottomrule
  \end{tabular}  
  \caption{Number of LLM families covered by different benchmarks. Many of the benchmarks are for factuality rather than faithfulness in this paper, or they do have have human ground truth.}
  \label{tab:benchmark_coverage}
\end{table*}

A team from University of Edinburgh~\cite{hugginface-hallucinations-leaderboard} evaluates LLMs' ability to serve as hallucination detectors, i.e., LLM-as-a-judge, on various tasks. The data may be human-written, LLM-generated, or rule-synthesized. In contrast, our work here focus on hallucinations in LLM-generated content. Another difference between  this work and our work is that this work checks both faithfulness and factuality while our work focuses on faithfulness because we believe that LLMs's value is more in their ability to respond to variable inputs than in answering questions 
using knowledge parametrized, or memorized. 

\section{Samples where LLMs hallucinate}
\label{sec:app:trick}

We found many samples that LLMs (are successfully tricked to) hallucinate. Here are a couple: 

\begin{itemize}
    \item By GPT-4o:\\  ``Hourglass'' is a song by British electronic duo Disclosure. Hourglass is singer-songwriter James Taylor's fourteenth studio album.\\ --> The passage describes that ``Hourglass'' is a song by the British electronic duo Disclosure. This song is featured on singer-songwriter James Taylor's fourteenth studio album.\\
    \textcolor{blue}{(The passage does not say whether the song was in the album. But the LLM linked them together.)}
    \item By Mistral-7B:\\ Brad Wilk. He is best known as the drummer of the rock bands Rage Against the Machine (1991--2000; 2007--11) and Audioslave (2001--07). Wilk started his career as a drummer for Greta in 1990, and helped co-found Rage with Tom Morello and Zack de la Rocha in August 1991. As of 2016, he is playing in the band Prophets of Rage, with Commerford, Morello, Chuck D, B-Real and DJ Lord. Rage is a German heavy metal band, formed in 1984 by Peter ``Peavy'' Wagner.\\ --> Brad Wilk is a drummer who gained prominence as a member of the rock bands Rage Against the Machine (1991-2000, 2007-2011) and Audioslave (2001-2007). He began his career with the band Greta in 1990 and co-founded Rage with Tom Morello and Zack de la Rocha in August 1991. As of 2016, he is playing in the band Prophets of Rage. Rage Against the Machine is not a German heavy metal band, as stated in the passage, but rather an American rock band. The passage incorrectly mentions that Rage is a German heavy metal band, formed in 1984 by Peter ``Peavy'' Wagner.\\
    \textcolor{blue}{(The LLM gets confused and thinks the info of the German band, which is factual and faithful, is a hallucination.)}
    \item By Qwen\\
    The "black box" of the Su-24 jet was officially opened in Moscow on Friday in front of journalists and diplomats. Nikolai Primak, head of the Russian investigation, said flight information appeared to be missing. \\ -> 
    The black box from the Su-24 jet was opened in Moscow, revealing potentially missing flight information. 
\end{itemize}

\section{AI assistant usage}
We used AI assistants in generating analytics code and revising the paper occasionally. 

\end{document}